\documentclass{article}
\usepackage{spdmetaformer,times}

\usepackage{amsmath,amsfonts}
\usepackage{microtype}
\usepackage{graphicx}
\usepackage{booktabs}
\usepackage{xurl}
\usepackage[colorlinks=true,linkcolor=blue,citecolor=black,urlcolor=blue]{hyperref}
\hypersetup{
  pdftitle={SPD-MetaFormer Is What You Need for Small-Data Brain Decoding},
  pdfauthor={Zhida Wang, Wei Lyu, Guo Yu, Sui Tang}
}

\title{SPD-MetaFormer Is What You Need for Small-Data Brain Decoding}

\author{Zhida Wang, Wei Lyu, Guo Yu, Sui Tang}

\begin{document}

\maketitle

\begin{abstract}
Brain signal decoding is challenging because neural recordings are noisy and vary across individuals, while labeled data are often limited. Recent attention-based models on the symmetric positive definite (SPD) manifold have nevertheless achieved strong performance using covariance and connectivity representations, yet the contribution of learned token weighting remains unclear. We examine two representative architectures, MAtt (based on log-Euclidean geometry) and GBWAtt (based on generalized Bures--Wasserstein geometry), and find that their learned attention weights remain close to uniform after training. We relate this behavior to bounded similarity parameterizations that, under the original softmax scaling, limit attention-weight contrast. Moreover, replacing learned weights with uniform weights, throughout training and evaluation, has little effect on mean predictive performance while preserving each model's original aggregation geometry. Motivated by these findings, we introduce SPD-MetaFormer, an attention-free architecture built on uniformly weighted Fr\'echet aggregation under log-Euclidean geometry. Its backbone uses a geodesic residual to update a summary token and a shared spectral feedforward map to transform all tokens, followed by a learned weighted readout. Token states remain SPD-valued until tangent-space classification. Across three EEG benchmarks, SPD-MetaFormer achieves competitive results relative to published Euclidean and manifold baselines. Separate matched reproductions test learned versus uniform weighting within MAtt and GBWAtt. These results suggest that, in the short-sequence and limited-data regimes studied, carefully designed SPD architectures can provide a simpler and effective alternative to adaptive manifold attention.
\end{abstract}

\section{Introduction}
\label{sec:introduction}

Understanding the human brain is a long-standing challenge in science and medicine. Brain decoding connects neural recordings to cognitive and clinical states, supporting brain--computer interfaces, neurorehabilitation, and the study of neurological and psychiatric conditions such as ADHD, Parkinson's disease, and depression \citep{gbd2024nervous,gbd2022mental,wolpaw2002bci,chaudhary2016bci,woo2017biomarkers}. The learning problem is demanding because recordings are high-dimensional, noisy, and heterogeneous, while labeled data are often limited and costly to collect \citep{lotte2018review,varoquaux2018crossvalidation}. Reliable decoding must therefore recover stable structure without overfitting to subject- and session-specific variation.

Existing approaches include convolutional decoders for electroencephalography (EEG) signals \citep{lawhern2018eegnet,mane2021fbcnet} and methods that explicitly model second-order structure. In EEG, covariance matrices capture power and coordinated channel variation relevant to motor imagery (MI), steady-state visual evoked potentials (SSVEPs), and error-related negativity (ERN), while functional magnetic resonance imaging (fMRI) connectivity describes covariation among regional time series \citep{barachant2012multiclass,congedo2017riemannian,dai2020analyzing,you2021revisiting}. Although these quantities have different neuroscientific meanings, appropriately regularized estimates from both are symmetric positive definite (SPD) matrices. Once a metric is specified, SPD geometry supplies meaningful distances and averages without learning these operations from scarce labels, supporting classical Riemannian classifiers and deep SPD networks \citep{pennec2006riemannian,arsigny2006log,huang2017riemannian,brooks2019riemannian}.

Building on these representations, recent methods use Transformer-style attention to model interactions among manifold-valued tokens. Manifold attention assigns these tokens input-dependent weights before geometry-aware aggregation. MAtt \citep{pan2022matt} and GBWAtt \citep{jin2026gbwatt} are representative examples, using Log--Euclidean and generalized Bures--Wasserstein geometries, respectively. Other methods explore deep SPD features, correlation matrices, and gyrovector spaces \citep{wang2023riemannian,hu2025coratt,wang2025gyroatt}. Related Transformers likewise preserve SPD structure throughout their layers \citep{seraphim2024structure}. Although these models achieve competitive performance, it remains unclear how much the learned attention weights contribute to the improvement.

We investigate this question in MAtt and GBWAtt by comparing learned attention weights with uniform weights while keeping the rest of each architecture unchanged. We find that both models learn nearly uniform attention weights. Replacing these weights with uniform weights during both training and evaluation has little effect on mean predictive performance, while preserving each model's original aggregation geometry (see Section~\ref{sec:uniform-diagnosis}; Appendix~\ref{app:uniform-protocol} and Appendix~\ref{app:temperature}). Our analysis shows that bounded scores limit the contrast between attention weights at unit temperature, while temperature controls reveal no consistent benefit from sharper attention. Together, these findings suggest that attention in these two architectures acts primarily as geometry-aware averaging in the studied regime. This motivates using uniform geometric aggregation as the token mixer and designing effective transformations around it.

MetaFormer, developed for computer vision tasks \citep{yu2022metaformer}, and related work on token mixing \citep{tay2021synthesizer} mirror these observations by demonstrating that simplified token mixers suffice for effective representations.
Adapting these ideas to small-data brain decoding requires operations that preserve SPD structure while combining information across covariance groups and keeping local evidence available for classification. We therefore introduce SPD-MetaFormer, which retains local covariance tokens and refines a separate summary through uniform geometric aggregation of transformed tokens. A shared spectral transformation improves matrix conditioning, and a geometric readout combines the local tokens and summary using learned weights shared across inputs. The resulting representation enters a tangent space only for final classification.

In summary, we make the following contributions:
\begin{itemize}
    \item \textbf{Controlled diagnosis of manifold attention.}
    Across three EEG benchmarks, MAtt and GBWAtt learn nearly uniform attention weights, and matched uniform-weight interventions change mean scores by at most $0.36$ points. Score analysis bounds attention-weight contrast at unit temperature, while temperature controls reveal no consistent benefit from sharper attention in the studied settings (Section~\ref{sec:uniform-diagnosis}; Appendix~\ref{app:temperature}).
    \item \textbf{SPD-MetaFormer.}
    We introduce an attention-free SPD backbone that separates local covariance memories from a global summary. Uniform Log--Euclidean aggregation of transformed memories drives a gated geodesic update of the summary, while keeping local tokens available for readout. A shared, single-parameter spectral transformation preserves positive definiteness and cannot increase the spectral condition number of any token. A geometric readout then combines local and global information using learned weights shared across inputs (Sections~\ref{sec:tokens}--\ref{sec:readout}).
    \item \textbf{Empirical evaluation.}
    SPD-MetaFormer achieves mean accuracies of $75.44\%$ and $70.79\%$ on MI and SSVEP and a mean AUC of $82.53\%$ on ERN, exceeding the highest baseline means in the published comparison compiled by \citet{jin2026gbwatt}. On ABIDE fMRI, it achieves $79.14\%$ AUC, the highest mean among the compared baselines under shared subject splits.
\end{itemize}

\section{Preliminaries}
\label{sec:prelim}

Let $\mathbb{S}^{d}$ and $\mathbb{S}_{++}^{d}$ denote the real symmetric
and symmetric positive-definite (SPD) $d\times d$ matrices, respectively.
For an eigendecomposition $A=U\operatorname{diag}(\lambda_1,\ldots,\lambda_d)U^{\top}$,
matrix functions act on the eigenvalues:
$f(A)=U\operatorname{diag}(f(\lambda_1),\ldots,f(\lambda_d))U^{\top}$.
In particular, $\log$ maps SPD matrices to $\mathbb{S}^{d}$, and $\exp$
maps symmetric matrices back to $\mathbb{S}_{++}^{d}$. A complementary
operation is the congruence map $X\mapsto WXW^{\top}$, which preserves
positive definiteness for full-row-rank $W\in\mathbb{R}^{r\times d}$
\citep{huang2017riemannian}.

For $X,Y\in\mathbb{S}_{++}^{d}$, the Log-Euclidean (LE) metric measures
distance in the matrix-logarithm coordinates \citep{arsigny2006log}:
\begin{equation}
d_{\mathrm{LE}}(X,Y)
=\left\lVert\log X-\log Y\right\rVert_{\mathrm{F}}.
\label{eq:lem}
\end{equation}
For $X_{1:m}=(X_1,\ldots,X_m)\in(\mathbb{S}_{++}^{d})^{m}$ and weights
$\boldsymbol{w}=(w_1,\ldots,w_m)$ with $w_j\geq 0$ and
$\sum_j w_j=1$, the weighted Fr\'echet mean uniquely minimizes
$\sum_{j=1}^{m}w_j d_{\mathrm{LE}}^{2}(P,X_j)$ over
$P\in\mathbb{S}_{++}^{d}$ and has the closed form
\citep{arsigny2006log,arsigny2007geometric}
\begin{equation}
\operatorname{FM}^{\mathrm{LE}}_{\boldsymbol{w}}(X_{1:m})
=\exp\!\left(\sum_{j=1}^{m}w_j\log X_j\right).
\label{eq:lem-closed-form}
\end{equation}
Similarly, the LE geodesic interpolates between $X,Y\in\mathbb{S}_{++}^{d}$
while remaining SPD:
\begin{equation}
X\mathbin{\#_{t}}Y
=\exp\!\left((1-t)\log X+t\log Y\right),
\qquad t\in[0,1].
\label{eq:geodesic}
\end{equation}

Section~\ref{sec:method} uses congruence maps for covariance embeddings and
mixer projections, spectral maps for token transformations, and $\log$ for
the final tangent representation. The mean
$\operatorname{FM}^{\mathrm{LE}}_{\boldsymbol{w}}$ supplies uniform mixing
and weighted readout, while $\#_{t}$ defines the gated summary residual.

\section{Proposed Method}
\label{sec:method}

\subsection{Architecture Overview and Design Rationale}
\label{sec:method-overview}

We introduce SPD-MetaFormer, a covariance-based decoder that uses uniform
geometric mixing directly on SPD matrices. The central design
problem is to integrate information across covariance groups and preserve the
evidence contributed by each group. SPD-MetaFormer addresses this problem with
local covariance memory tokens that retain group-specific second-order
statistics and a data-dependent summary that accumulates global information.
This role separation keeps the local sources accessible throughout summary
refinement and final readout.

\begin{figure}[t]
  \centering
  \includegraphics[width=\linewidth]
  {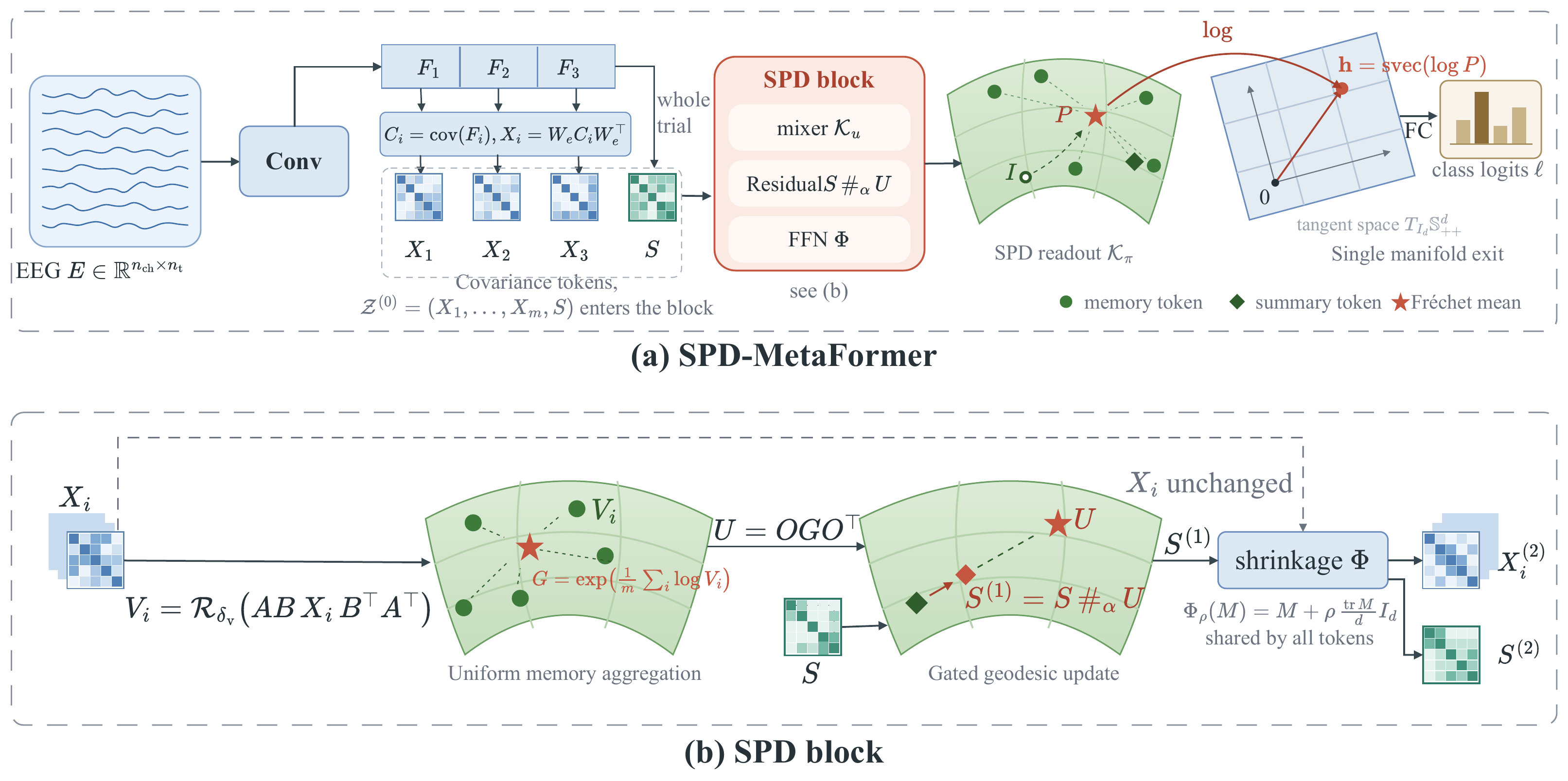}
  \caption{Overview of SPD-MetaFormer.
  \textbf{(a)}~Convolutional features yield local covariance memories (blue)
  and an observed summary (green), embedded in a common SPD space.
  After the SPD block, a geometric readout with learned weights shared across
  trials (dotted spokes) pools all tokens for tangent-space classification.
  \textbf{(b)}~The attention-free block updates only the summary through
  uniform geometric memory aggregation and a gated geodesic residual, then
  applies shared spectral shrinkage to every token.
  Circles, diamonds, and stars denote memories, the summary, and
  Fr\'echet means, respectively.}
  \label{fig:spd-metaformer}
\end{figure}

Following the MetaFormer perspective \citep{yu2022metaformer},
Figure~\ref{fig:spd-metaformer} separates geometric token mixing from shared
spectral shrinkage, followed by geometric readout and tangent-space classification.

\subsection{Covariance Tokens Separate Local Evidence from a Global State}
\label{sec:tokens}

The left side of Figure~\ref{fig:spd-metaformer}(a) constructs the role-aware
input from the same trial at local and global resolutions.

\paragraph{Local covariance memory.}
The memory path represents an EEG trial through a collection of local
second-order statistics. Given
$E\in\mathbb{R}^{n_{\mathrm{ch}}\times n_{\mathrm t}}$ with
$n_{\mathrm{ch}}$ channels and $n_{\mathrm t}$ time samples, spatial and temporal
convolutions extract a feature map $F=f_{\mathrm{conv}}(E)$. Following the
spatial--temporal decomposition used in EEG convolutional models and manifold
attention \citep{wei2019sccnet,pan2022matt,jin2026gbwatt}, we divide $F$ into
$m$ feature groups $F_i\in\mathbb{R}^{c_{\mathrm f}\times L_i}$ for $i = 1, \ldots, m$, each with
$c_{\mathrm f}$ feature channels and $L_i$ time samples, and convert each group
into a regularized covariance,
\begin{equation}
 C_i=\frac{1}{L_i-1}\bar F_i\bar F_i^{\top}+\varepsilon I_{c_{\mathrm f}},
 \qquad
 \bar F_i=F_i-\frac{1}{L_i}F_i\mathbf{1}_{L_i}\mathbf{1}_{L_i}^{\top}.
 \label{eq:cov-tokens}
\end{equation}
Here $I_{c_{\mathrm f}}$ is the $c_{\mathrm f}\times c_{\mathrm f}$ identity matrix and
$\mathbf{1}_{L_i}$ is the $L_i$-dimensional all-ones vector. We set the
diagonal floor to $\varepsilon=10^{-5}$ to ensure positive definiteness and
stabilize subsequent matrix operations. We preserve the original covariance
scale, so each $C_i$ records both the second-order structure and the energy of
its feature group.

\paragraph{Observed global state.}
The summary construction follows the feature-grouping scheme. For temporal
segments, $C_{\mathrm{s}}$ is computed from the complete feature sequence. For
channel groups that each span the full trial, we aggregate their covariances as
$C_{\mathrm{s}}=m^{-1}\sum_{i=1}^{m}C_i$. The resulting $C_{\mathrm{s}}$
initializes the trial-specific state updated by the memory mixer. The
construction used for each task is specified with the corresponding data
preprocessing.

\paragraph{Common SPD space.}
Geometric interaction requires placing local and global states in a common space. We
therefore apply a shared congruence map
\citep{huang2017riemannian},
\begin{equation}
 X_i=W_{\mathrm e} C_i W_{\mathrm e}^{\top},\qquad
 S=W_{\mathrm e} C_{\mathrm s} W_{\mathrm e}^{\top},\qquad
 W_{\mathrm e}\in\mathbb{R}^{d\times c_{\mathrm f}},\quad
 W_{\mathrm e}W_{\mathrm e}^{\top}=I_d.
 \label{eq:input-bimap}
\end{equation}
Row orthonormality keeps the map full rank and hence preserves positive
definiteness. When $c_{\mathrm f}=d$, we omit this map and use the identity. The
result is the role-aware sequence
$\mathcal Z^{(0)}=(X_1,\ldots,X_m,S)$ shown at the entrance to the SPD block in
Figure~\ref{fig:spd-metaformer}. Across datasets, the model retains this token
structure and adapts the number of memory tokens and their SPD dimension
to the input feature structure. Appendix~\ref{app:implementation} specifies the
exact configurations, feature shapes, and numerical eigenvalue floors. With
both roles expressed in the same geometry, they form a valid input to the
geometric mixer.

\subsection{Uniform Geometric Mixing Refines the Summary}
\label{sec:mixer}
Uniform geometric mixing is motivated by the near-uniform allocations learned
by MAtt and GBWAtt. This behavior suggests that their effective operation is
geometric averaging rather than selective routing. We therefore retain learned
SPD transformations and Fr\'echet aggregation while replacing input-dependent
weights with uniform mixing. Figure~\ref{fig:spd-metaformer}(b) illustrates this
design. Its empirical basis is examined in
Section~\ref{sec:uniform-diagnosis}, with further analyses in
Appendix~\ref{app:uniform-protocol} and Appendix~\ref{app:temperature}.

\paragraph{Uniform memory aggregation.}
Let $B\in\mathbb{R}^{d\times d}$ be a learned orthogonal base map, and let
$A,O\in\mathbb{R}^{d\times d}$ be learned value and output maps, respectively.
Using eigenvalue rectification $\mathcal{R}_{\delta}$ as in SPDNet
\citep{huang2017riemannian}, the mixer computes
\begin{equation}
 V_i=\mathcal{R}_{\delta_v}
       \bigl(A B X_i B^{\top}A^{\top}\bigr).
 \label{eq:uniform-mixer}
\end{equation}
The transformed memory tokens have the uniform LE mean
\begin{equation}
 G=\operatorname{FM}^{\mathrm{LE}}_{\frac{1}{m}\mathbf{1}_m}
      (\{V_i\}_{i=1}^{m})
   =\exp\!\left(\frac{1}{m}\sum_{i=1}^{m}\log V_i\right).
\end{equation}
The output map gives $U=OGO^{\top}$.
We set $\delta_v=10^{-4}$ to floor the eigenvalues of $V_i$, preserving
positive definiteness and stabilizing the matrix logarithms. The intermediate
$G$ is the uniform LE Fr\'echet mean from
Equation~\ref{eq:lem-closed-form} \citep{arsigny2006log,arsigny2007geometric},
and $U$ is the transformed memory mean passed to the gated summary update.

\paragraph{Gated summary update.}
The transformed memory mean $U$ supplies cross-group information to the
summary. We interpolate $S$ and $U$ along an LE geodesic,
\begin{equation}
 S^{(1)}=S\mathbin{\#_{\alpha}}U
 =\exp\!\left((1-\alpha)\log S+\alpha\log U\right),
 \qquad
 \alpha=\operatorname{sigmoid}(a),
 \label{eq:summary-update}
\end{equation}
where $a$ is learned. The gate controls
the update strength, and LE interpolation preserves positive definiteness.
In parallel, the memory tokens follow the identity path $X_i^{(1)}=X_i$, yielding
$\mathcal Z^{(1)}=(X_1^{(1)},\ldots,X_m^{(1)},S^{(1)})$. This separation
parallels the distinct feature and aggregation roles in class-attention
architectures \citep{touvron2021cait}. The shared spectral shrinkage is then
applied to both paths.

\subsection{Shared Spectral Shrinkage Conditions Both Token Paths}
\label{sec:spectral-ffn}

The mixer deliberately gives the memory and summary different update
histories. Before readout, the feedforward part of the SPD block applies the
same spectral shrinkage to both, improving their conditioning and preserving
their roles. For $M\in\mathbb{S}_{++}^{d}$, define
\begin{equation}
 \Phi_{\rho}(M)
 =M+\rho\,\frac{\operatorname{tr}(M)}{d}I_d,
 \qquad
 \rho=0.1\exp(\theta),
 \label{eq:spectral-ffn}
\end{equation}
where the scalar $\theta$ is learned, initialized at zero, and shared across
tokens. Applying this map to the two paths gives
\begin{equation}
 X_i^{(2)}=\Phi_\rho(X_i^{(1)}),\qquad
 S^{(2)}=\Phi_\rho(S^{(1)}).
 \label{eq:block-output}
\end{equation}
The shared parameter $\rho$ yields a token-adaptive correction whose magnitude
scales with each token's mean eigenvalue.

The action of $\Phi_\rho$ on the eigenvalues is explicit. If $M$ has
eigenvalues $\lambda_k$ and mean eigenvalue
$\bar\lambda=d^{-1}\sum_k\lambda_k$, then $\Phi_\rho(M)$ retains the same
eigenvectors and has eigenvalues $\lambda_k+\rho\bar\lambda$. Positive
definiteness is therefore preserved, and the spectral condition number cannot
increase. Up to the retained scale factor $1+\rho$, this operation is a convex
combination of $M$ and its isotropic component
$\operatorname{tr}(M)I_d/d$, connecting it to covariance shrinkage
\citep{ledoit2004well}. The decoding objective directly optimizes $\rho$, which
distinguishes it from a covariance-estimation coefficient. This token-wise
operation complements the preceding cross-token mixer.

\subsection{Trial-Independent Geometric Readout Retains Local and Global Evidence}
\label{sec:readout}

After summary refinement and spectral shrinkage, the block returns
$\mathcal Z^{(2)}=(X_1^{(2)},\ldots,X_m^{(2)},S^{(2)})$. The readout in
Figure~\ref{fig:spd-metaformer}(a) fuses every token on the SPD manifold before
the architecture makes its single transition to a Euclidean decision space.

\paragraph{Trial-independent geometric readout.}
The readout assigns one learned score $b_i$ to each memory position and one
score $\beta$ to the summary. These scores are shared across trials and give
the normalized role weights
\begin{equation}
 \boldsymbol{\pi}
 =\frac{\bigl(\exp(b_1),\ldots,\exp(b_m),\exp(\beta)\bigr)}
 {\sum_{j=1}^{m}\exp(b_j)+\exp(\beta)}.
 \label{eq:readout-weights}
\end{equation}
The readout $\mathcal K_\pi$ then forms their weighted LE Fr\'echet mean,
\begin{equation}
 P=\exp\!\left(
 \sum_{i=1}^{m}\pi_i\log X_i^{(2)}
 +\pi_{m+1}\log S^{(2)}
 \right).
 \label{eq:geometric-readout}
\end{equation}
The mixer uses uniform weights for the gated summary update. The readout
learns role weights for the final fusion of the memories and summary.

\paragraph{Single manifold exit.}
After geometric readout, the single SPD representation $P$ is mapped to the
tangent space at the identity and classified with one affine layer,
\begin{equation}
 \mathbf h=\operatorname{svec}(\log P),\qquad
 \widehat{\mathbf p}(\cdot\mid E)
 =\operatorname{softmax}(W_h\mathbf h+b_h).
 \label{eq:classifier}
\end{equation}
Here $\mathbf{h}\in\mathbb{R}^{d(d+1)/2}$, and
$\widehat p(y\mid E)=\widehat{\mathbf p}_y$. The operator
$\operatorname{svec}$ stacks the upper triangle and multiplies off-diagonal
entries by $\sqrt{2}$, preserving the Frobenius inner product on symmetric
matrices. This tangent head follows established SPD networks
\citep{huang2017riemannian,pan2022matt} and is applied once, after the manifold
backbone has completed summary refinement and sequence readout. All modules are
trained jointly with cross-entropy, with class weighting for the imbalanced ERN
task.

\section{Experiments}
\label{sec:experiments}

We evaluate SPD-MetaFormer on three public EEG benchmarks spanning
motor imagery (MI), steady-state visual evoked potentials (SSVEP), and
error-related negativity (ERN). An additional study on the ABIDE fMRI
cohort examines whether the same SPD processing architecture transfers
beyond EEG, with the complete protocol and results reported in
Appendix~\ref{app:fmri}. The evaluation first establishes its
predictive performance against existing convolutional and geometry-aware
EEG models. Controlled comparisons then examine the contribution of the
complete SPD processing path and its static geometric readout. Finally,
we revisit MAtt and GBWAtt through locked-uniform interventions and
allocation-entropy measurements to determine whether their
content-dependent weights contribute to predictive performance.

\paragraph{Datasets and protocol.}
We use BCIC-IV-2a for four-class MI decoding with nine subjects
\citep{brunner2008bci}, MAMEM-SSVEP-II for five-class SSVEP decoding
with eleven subjects \citep{nikolopoulos2016mamem}, and BCI-ERN for
binary ERN detection with sixteen subjects \citep{perrin2012objective}, following prior
manifold-attention studies \citep{pan2022matt,jin2026gbwatt}.
Each model is trained independently for each subject using seeds $0$--$9$.
For MI, we form a
class-stratified validation subset from the training session and use the
separate evaluation session for testing. For SSVEP and ERN, the first four
sessions support training and validation (one held out), and the final
session is used for testing, following MAtt \citep{pan2022matt}.
All paired ablations share the data splits and select the checkpoint with
the lowest validation loss. Full training settings appear in
Appendix~\ref{app:protocol}.

\paragraph{Metrics and uncertainty.}
We report accuracy for MI/SSVEP and ROC-AUC for imbalanced ERN, on a
0--100 scale. ERN AUC follows the evaluation convention of MAtt and GBWAtt
\citep{pan2022matt,jin2026gbwatt}. Our scores average seeds within subjects, then
subjects equally; $\pm$ values and error bars report the mean within-subject
sample SD across runs, also used for entropy dispersion.
Paired bootstrap intervals resample subjects after seed averaging and are
interpreted descriptively without multiplicity correction
(Appendix~\ref{app:protocol}). Published baselines retain their source
uncertainty definitions.

\subsection{Comparison with Existing Methods}
\label{sec:benchmark}

\begin{table}[t]
\centering
\caption{Comparison with published EEG results. MI and SSVEP report accuracy (\%); ERN reports AUC (\%). Baseline entries are reproduced from Table~3 of \citet{jin2026gbwatt}.}
\label{tab:benchmark}
\setlength{\tabcolsep}{5pt}
\begin{tabular}{@{}lccc@{}}
\toprule
Model & MI & SSVEP & ERN \\
\midrule
\multicolumn{4}{l}{\emph{Published results: mean $\pm$ SD}} \\
EEGNet & $61.84 \pm 6.39$ & $53.72 \pm 7.23$ & $74.28 \pm 2.47$ \\
ShallowCNet & $57.43 \pm 6.25$ & $56.93 \pm 6.97$ & $71.86 \pm 2.64$ \\
SCCNet & $71.95 \pm 5.05$ & $62.11 \pm 7.70$ & $70.93 \pm 2.31$ \\
FBCNet & $56.52 \pm 3.07$ & $53.09 \pm 5.67$ & $60.47 \pm 3.06$ \\
TCNet-Fusion & $71.45 \pm 4.45$ & $45.00 \pm 6.45$ & $70.46 \pm 2.94$ \\
MBEEGSE & $64.58 \pm 6.07$ & $56.45 \pm 7.27$ & $75.46 \pm 2.34$ \\
SPDNet & $72.93 \pm 4.33$ & $62.30 \pm 3.12$ & $72.05 \pm 4.43$ \\
SPDNetBN & $73.02 \pm 3.67$ & $62.76 \pm 3.01$ & $72.34 \pm 3.46$ \\
MAtt & $74.71 \pm 5.01$ & $65.19 \pm 3.14$ & $75.68 \pm 2.23$ \\
GDLNet & $69.32 \pm 2.89$ & $65.52 \pm 2.86$ & $78.23 \pm 2.52$ \\
GBWAtt ($\theta=1$) & $74.54 \pm 2.04$ & $67.12 \pm 2.94$ & $80.42 \pm 2.12$ \\
GBWAtt ($\theta=1.5$) & $74.95 \pm 1.83$ & $67.27 \pm 2.62$ & $80.90 \pm 1.97$ \\
\midrule
\multicolumn{4}{l}{\emph{Our experiments: mean $\pm$ SD}} \\
\textbf{SPD-MetaFormer} & $\mathbf{75.44 \pm 1.69}$ & $\mathbf{70.79 \pm 2.17}$ & $\mathbf{82.53 \pm 1.85}$ \\
\bottomrule
\end{tabular}
\end{table}

We compare SPD-MetaFormer with convolutional EEG models, including
EEGNet \citep{lawhern2018eegnet}, ShallowCNet
\citep{schirrmeister2017deep}, SCCNet \citep{wei2019sccnet}, FBCNet
\citep{mane2021fbcnet}, TCNet-Fusion \citep{musallam2021tcnet}, and
MBEEGSE \citep{altuwaijri2022mbeegse}. The geometry-aware baselines are
SPDNet \citep{huang2017riemannian}, SPDNetBN
\citep{brooks2019riemannian}, MAtt \mbox{\citep{pan2022matt}}, GDLNet
\citep{wang2024gdlnet}, and GBWAtt \citep{jin2026gbwatt}. Their results
are taken from the published comparison of \citet{jin2026gbwatt} rather
than from new reproductions.

SPD-MetaFormer achieves $75.44\%$ accuracy on MI, $70.79\%$ on SSVEP,
and $82.53\%$ AUC on ERN. These results exceed the highest published
mean in each column by $0.49$, $3.52$, and $1.63$ points, respectively.
The largest improvement occurs on SSVEP, whereas the MI result remains
close to the strongest reported baseline. The consistent performance
across the three paradigms shows that an attention-free architecture can
remain competitive when exact uniform LE mixing is combined with the
shared spectral transformation and static geometric readout.

To examine whether the proposed SPD processing path transfers beyond EEG,
we further evaluate SPD-MetaFormer on the multi-site ABIDE fMRI cohort.
It achieves an AUC of $79.14\%$ and exceeds the strongest compared baseline under matched subject splits. This result provides complementary evidence that the same geometry-aware architecture transfers
to covariance representations from a distinct neuroimaging modality.
The complete protocol and results are reported in
Appendix~\ref{app:fmri}.

\subsection{Whole-Path Comparisons}
\label{sec:whole-path}

\begin{table}[t]
\centering
\caption{Whole-path controls under the same subject/seed grid and training recipe. Values are mean $\pm$ mean within-subject run SD (sample SD over ten runs per subject, averaged over subjects). MI/SSVEP: accuracy (\%); ERN: AUC (\%).}
\label{tab:whole-path}
\setlength{\tabcolsep}{5pt}
\begin{tabular}{@{}lccc@{}}
\toprule
Configuration & MI & SSVEP & ERN \\
\midrule
Convolutional features only & $26.25 \pm 2.03$ & $24.35 \pm 3.15$ & $72.29 \pm 3.71$ \\
SPD tangent readout & $70.27 \pm 2.26$ & $59.35 \pm 3.10$ & $80.87 \pm 2.49$ \\
Euclidean MetaFormer & $59.32 \pm 3.94$ & $67.69 \pm 3.06$ & $77.65 \pm 3.29$ \\
Memory-only geometric readout & $74.41 \pm 1.90$ & $67.93 \pm 3.15$ & $80.83 \pm 2.50$ \\
\midrule
\textbf{Full SPD-MetaFormer} & $\mathbf{75.44 \pm 1.69}$ & $\mathbf{70.79 \pm 2.17}$ & $\mathbf{82.53 \pm 1.85}$ \\
\bottomrule
\end{tabular}
\end{table}

The benchmark results do not determine whether performance arises from
the convolutional frontend, covariance tokenization alone, or the
complete SPD block and static geometric readout. We therefore construct
four controls that share the frontend initialization, subject--seed
grid, and optimization recipe of SPD-MetaFormer.
\emph{Convolutional features only} flattens the feature map
$F=f_{\mathrm{conv}}(E)$ and applies a linear classifier, removing
covariance tokenization and all subsequent SPD operations.
\emph{SPD tangent readout} retains the covariance tokens, data-dependent
summary, and shared congruence map $W_{\mathrm e}$, but replaces the SPD
block and static geometric readout with a linear classifier applied to
the concatenated log-vectorizations of
$\mathcal Z^{(0)}=(X_1,\ldots,X_m,S)$. \emph{Euclidean MetaFormer}
preserves the same covariance tokenization and shared congruence map,
then processes the log-vectorized tokens using a pre-LayerNorm uniform
mixer, additive residuals, and a two-layer GELU MLP with an expansion
ratio of four, followed by mean readout.
\emph{Memory-only geometric readout} removes the summary and the SPD
block, applying the static geometric readout
(equation~\ref{eq:geometric-readout}) to $X_1,\ldots,X_m$ alone.

SPD-MetaFormer outperforms all four controls on every dataset
(Table~\ref{tab:whole-path}). Relative to direct SPD tangent readout, it
improves MI and SSVEP accuracy by $5.17$ and $11.44$ points and ERN AUC
by $1.66$ points. The Euclidean MetaFormer also performs worse despite
having more parameters (Appendix~\ref{app:ablation-details}). Retaining
the geometric readout but removing the summary and SPD block still
lowers MI, SSVEP, and ERN by $1.03$, $2.86$, and $1.70$ points. The
paired subject-bootstrap intervals exclude zero in all twelve comparisons.
The control hierarchy argues against four simpler explanations for the
observed performance. The convolutional frontend alone is insufficient,
constructing SPD tokens followed by direct tangent classification is
insufficient, geometric readout of the local memories alone is
insufficient, and generic uniform mixing in a Euclidean backbone does not
recover the full model. These results support the complete SPD processing
path rather than uniform mixing alone. Since the controls jointly alter
transformations, capacity, and readout, the next experiment more narrowly
isolates the static geometric readout.

\subsection{Readout Ablations}
\label{sec:readout-ablation}

\begin{table}[t]
\centering
\caption{Readout ablations with the complete SPD block retained. All three controls change the final representation used for classification. Values are mean $\pm$ mean within-subject run SD (sample SD over ten runs per subject, averaged over subjects). MI/SSVEP: accuracy (\%); ERN: AUC (\%).}
\label{tab:readout-ablation}
\setlength{\tabcolsep}{5pt}
\begin{tabular}{@{}lccc@{}}
\toprule
Configuration & MI & SSVEP & ERN \\
\midrule
Concatenation readout & $70.31 \pm 2.45$ & $61.73 \pm 3.31$ & $81.83 \pm 1.93$ \\
Arithmetic pooling & $75.31 \pm 1.86$ & $66.65 \pm 2.35$ & $81.90 \pm 2.00$ \\
Summary-only readout & $74.06 \pm 1.85$ & $67.03 \pm 2.94$ & $82.06 \pm 2.04$ \\
\midrule
\textbf{Full SPD-MetaFormer} & $\mathbf{75.44 \pm 1.69}$ & $\mathbf{70.79 \pm 2.17}$ & $\mathbf{82.53 \pm 1.85}$ \\
\bottomrule
\end{tabular}
\end{table}

We retain the complete SPD block and modify only its static geometric
readout. \emph{Concatenation readout} stacks
$\operatorname{svec}(\log X_i^{(2)})$ for all memory tokens together with
$\operatorname{svec}(\log S^{(2)})$ and learns a new linear classifier.
\emph{Arithmetic pooling} replaces the weighted LE Fr\'echet mean in
equation~\ref{eq:geometric-readout} with
$P_{\mathrm{arith}}=\sum_{i=1}^{m}\pi_iX_i^{(2)}+
\pi_{m+1}S^{(2)}$ while retaining the input-independent role weights
$\boldsymbol{\pi}$. \emph{Summary-only readout} discards the memory tokens
and applies the original affine classifier to
$\operatorname{svec}(\log S^{(2)})$.

The ablations reveal complementary roles for token retention and pooling
geometry. Concatenation retains every token but removes the structured
geometric pooling operation, producing the largest losses on MI and
SSVEP at $5.13$ and $9.06$ points, respectively. Summary-only readout
lowers the mean on all three tasks, indicating that the memory tokens
retain information that complements the refined summary at
classification. The effect of pooling geometry is more task-dependent.
Arithmetic pooling reduces SSVEP accuracy by $4.15$ points but changes MI
accuracy by only $0.13$ points. Overall, the static geometric readout
provides the most consistent performance.

\subsection{Locked-Uniform Attention and Allocation Entropy}
\label{sec:uniform-diagnosis}

\begin{table}[t]
\centering
\caption{Training with learned versus locked-uniform weights in existing MAtt and GBWAtt reproductions. Values are mean $\pm$ mean within-subject run SD (sample SD over ten runs per subject, averaged over subjects); $\Delta$ is uniform minus learned, with a paired subject-bootstrap 95\% interval. All scores are multiplied by 100. GBWAtt comparisons match $\theta=1.5$ on both sides. These experiments use their own matched reproduction protocols (Appendix~\ref{app:uniform-protocol}).}
\label{tab:uniform}
\setlength{\tabcolsep}{5pt}
\begin{tabular}{@{}lccc@{}}
\toprule
Model / dataset & Learned & Uniform & $\Delta$ [95\% interval] \\
\midrule
MAtt / MI & $74.84 \pm 1.69$ & $74.83 \pm 1.73$ & $-0.01\;[-0.21, +0.17]$ \\
MAtt / SSVEP & $65.14 \pm 3.33$ & $64.91 \pm 3.04$ & $-0.23\;[-1.22, +0.87]$ \\
MAtt / ERN & $74.87 \pm 4.41$ & $74.91 \pm 4.60$ & $+0.04\;[-1.15, +1.23]$ \\
GBWAtt ($\theta=1.5$) / MI & $75.21 \pm 1.57$ & $74.93 \pm 1.94$ & $-0.28\;[-0.99, +0.31]$ \\
GBWAtt ($\theta=1.5$) / SSVEP & $64.43 \pm 3.42$ & $64.06 \pm 2.87$ & $-0.36\;[-1.06, +0.35]$ \\
GBWAtt ($\theta=1.5$) / ERN & $80.97 \pm 2.35$ & $80.90 \pm 2.47$ & $-0.07\;[-0.42, +0.26]$ \\
\bottomrule
\end{tabular}
\end{table}

\begin{table}[t]
\centering
\caption{Attention entropy at the validation-selected checkpoint, averaged over test queries/trials, then seeds and subjects. $\overline{\mathrm{SD}}(H)$ averages the within-subject sample SD over ten runs. Natural logarithms give nats. Uniform weights have $H=\log m$ and $H/\log m=1$ exactly. GBWAtt uses $\theta=1.5$.}
\label{tab:entropy}
\setlength{\tabcolsep}{5pt}
\begin{tabular}{@{}lccccc@{}}
\toprule
Model / dataset & $m$ & $H_{\rm learned}$ & $\overline{\mathrm{SD}}(H)$ & $H_{\rm uniform}$ & $H_{\rm learned}/\log m$ \\
\midrule
MAtt / MI & 3 & $1.098605$ & $8.12\!\times\!10^{-7}$ & $1.098612$ & $0.999994$ \\
MAtt / SSVEP & 7 & $1.945906$ & $8.97\!\times\!10^{-7}$ & $1.945910$ & $0.999998$ \\
MAtt / ERN & 3 & $1.098578$ & $1.98\!\times\!10^{-5}$ & $1.098612$ & $0.999969$ \\
GBWAtt / MI & 3 & $1.098410$ & $3.44\!\times\!10^{-5}$ & $1.098612$ & $0.999815$ \\
GBWAtt / SSVEP & 7 & $1.945243$ & $1.33\!\times\!10^{-4}$ & $1.945910$ & $0.999657$ \\
GBWAtt / ERN & 3 & $1.097353$ & $6.41\!\times\!10^{-4}$ & $1.098612$ & $0.998854$ \\
\bottomrule
\end{tabular}
\end{table}

We revisit ten-seed experiments using MAtt and the released
implementation of GBWAtt. Within each architecture, the locked-uniform
control fixes every attention row to
$\mathbf u=m^{-1}\mathbf 1_m$ during both training and evaluation while
preserving the frontend, value transformation, aggregation geometry, and
classifier. MAtt retains its LE aggregation, whereas GBWAtt retains the
weighted square-root aggregation in its released implementation. We
denote the GBWAtt parameter by $\theta_{\mathrm{GBW}}$ and set it to
$1.5$ for both variants in each pair. The evaluation contains 720 matched
subject--seed pairs across the two architectures and three datasets, and Appendix~\ref{app:uniform-protocol} details the reproduction protocols.

Locking the attention weights changes mean performance by at most $0.36$
points across all six architecture--dataset comparisons
(Table~\ref{tab:uniform}), and every paired interval includes zero. Under
these configurations, the residual content-dependent variation in the
learned weights therefore provides no measurable improvement over
uniform allocation. The intervals are not equivalence tests, so the
result establishes practical insensitivity at the measured precision
rather than an exact zero effect.

We next determine whether this performance insensitivity coincides with
nearly uniform learned allocations. For a row-stochastic attention
matrix $\Gamma(Y)\in\mathbb{R}^{m\times m}$, we measure its entropy on the
test set at the selected checkpoint.
\begin{equation}
 H(\Gamma)=-\frac{1}{m}\sum_{i=1}^{m}\sum_{j=1}^{m}
 \Gamma_{ij}\log \Gamma_{ij},\qquad
 \overline H(\Gamma)=\frac{H(\Gamma)}{\log m}.
 \label{eq:attention-entropy}
\end{equation}
Uniform allocation gives $H(\Gamma)=\log m$ and
$\overline H(\Gamma)=1$. Learned MAtt attains a mean normalized entropy
above $0.99996$ on every dataset, while GBWAtt remains above $0.99885$
(Table~\ref{tab:entropy}). Both architectures therefore allocate nearly
equal mass across their short token sequences. The entropy measurement
characterizes the allocations learned by the original models, while the
locked-uniform intervention shows that removing their residual input
dependence has little effect on mean predictive performance.

The inverse-log score used by both implementations further constrains
the attainable allocation contrast \citep{pan2022matt,jin2026gbwatt}.
For a nonnegative distance $d_{ij}$, the score
$s_{ij}=1/(1+\log(1+d_{ij}))$ lies in $(0,1]$. At unit softmax
temperature, any two weights in the same row satisfy
$\Gamma_{ij}/\Gamma_{ik}=\exp(s_{ij}-s_{ik})\leq e$. The score
parameterization therefore bounds the pairwise allocation ratio
independently of the input. This structural constraint alone does not
imply the near-maximal empirical entropies. Temperature controls further
show that sharper attention does not consistently improve performance over
uniform mixing (Appendix~\ref{app:temperature}). Together, the intervention,
entropy measurements, and score analysis show that MAtt and GBWAtt
operate in a nearly uniform regime in which content-dependent weighting
has little effect on predictive performance. Combined with the
whole-path and readout controls, this diagnosis supports the design
emphasis of SPD-MetaFormer on uniform geometric mixing, shared spectral
transformation, and static geometric readout.

\section{Conclusion}
\label{sec:conclusion}

In this work, we revisited the role of adaptive manifold attention in small-data brain decoding. Matched experiments on MAtt and GBWAtt show that their learned allocations remain close to uniform and that replacing them with locked-uniform weights has little effect on predictive performance in the settings studied. Motivated by this finding, we introduced SPD-MetaFormer, an attention-free architecture built around uniform geometric mixing, geodesic summary refinement, shared spectral transformation, and geometric readout. Across three EEG benchmarks, SPD-MetaFormer achieves competitive performance relative to existing Euclidean and manifold-based methods, while additional results on ABIDE support its applicability to fMRI connectivity.
Overall, our results suggest that, for short sequences and limited data, the effectiveness of SPD models can arise from geometry-aware representation, transformation, and aggregation without requiring strongly selective input-dependent weighting. An important direction for future work is to understand why optimization nevertheless drives existing manifold-attention models toward nearly uniform allocations. Although bounded similarity scores constrain attainable attention contrast, they do not fully explain the near-maximal entropies observed in practice. A theoretical characterization of how score parameterization, optimization dynamics, data scale, and SPD geometry interact may help determine when adaptive manifold attention is necessary and when uniform geometric aggregation is sufficient.

\section*{AI Use Statement}
Generative AI tools were used to assist with brainstorming and refining research ideas, generating portions of the experimental code, and improving the clarity and readability of the manuscript. All AI-assisted ideas, code, and text were reviewed, verified, and, where necessary, modified by the authors. The authors take full responsibility for the final methodology, implementation, results, and content of the paper.

\section*{Reproducibility Statement}
Appendix~\ref{app:implementation} describes the feature extraction, covariance construction, parameterization, data splits, and training protocols. All datasets used in this work are publicly available.

\clearpage
\appendix
\section{Implementation and Evaluation Details}
\label{app:implementation}

\subsection{Feature Extraction and Covariance Construction}

Table~\ref{tab:frontend-details} specifies the frontend used for the
reported SPD-MetaFormer results. Every convolution is followed by
ordinary batch normalization. There is no additional activation between
these two Conv--BN stages. Temporal padding with an even kernel gives
one additional feature sample. Covariances are centered, divided by
$L_i-1$, and regularized by $10^{-5}I$. Neither trace normalization nor
an additional dimension-dependent multiplier is applied.

\begin{table}[htbp]
\centering
\caption{Dataset-specific SPD-MetaFormer frontend. $C_{\rm in}\!\to\!C_{\rm out}$ denotes convolutional feature channels, with kernel sizes shown in parentheses.}
\label{tab:frontend-details}
\small
\setlength{\tabcolsep}{4pt}
\begin{tabular}{@{}lccc@{}}
\toprule
 & MI & SSVEP & ERN \\
\midrule
Input dimensions & $22\times438$ & $8\times125$ & $56\times160$ \\
Spatial convolution & $1\!\to\!22\;(22,1)$ & $1\!\to\!125\;(8,1)$ & $3\!\to\!69\;(56,1)$ \\
Temporal convolution & $22\!\to\!20\;(1,12)$ & $125\!\to\!15\;(1,36)$ & $69\!\to\!48\;(1,64)$ \\
Temporal padding & 6 & 18 & 32 \\
Feature matrix & $20\times439$ & $15\times126$ & $48\times161$ \\
Memory grouping & time & time & channel banks \\
Memory group sizes & $147,146,146$ & $7\times18$ & $3\times16$ channels \\
Memory covariance dimension & 20 & 15 & 16 \\
Summary & full-trial cov. & full-trial cov. & bank-cov. mean \\
Entry map & $20\!\to\!15$ & identity & $16\!\to\!8$ \\
Memory + summary tokens & $3+1$ & $7+1$ & $3+1$ \\
\bottomrule
\end{tabular}
\end{table}

For ERN, the input is repeated into three streams and both convolutions
use three groups, yielding three separately learned banks of sixteen
channels. Each bank covariance uses all 161 output samples.
The reported ERN model uses the arithmetic summary.

\subsection{Parameterization and Numerical Operations}

Entry maps have orthonormal rows and the square base map is orthogonal.
They are parameterized through the thin QR decomposition of the
transpose of an unconstrained matrix, with diagonal-sign correction.
The value map is initialized as $A=I+0.1E/\sqrt d$ and the output map as
$O=I+0.02E'$, with independent standard-normal entries in $E,E'$.
These two square matrices are unconstrained during optimization.
Initialization near the identity is not a hard full-rank constraint.

The implementation symmetrizes matrix outputs. It applies an eigenvalue
floor of $10^{-6}$ before the entry map and $10^{-4}$ after the value
map. In float32, every matrix logarithm evaluates
$\log\max(\lambda,10^{-6})$ spectrally; this also protects the logarithm
of the output congruence in Equation~\ref{eq:summary-update}.
Thus, the exact SPD formulas in the main text have explicit numerical
regularization at their logarithms. Matrix exponentials and spectral
functions use eigendecompositions; backpropagation uses divided
differences with their repeated-eigenvalue limit.
All reported runs use float32 with two CPU threads per process.

The mixer changes only the summary before the spectral map, but
Equation~\ref{eq:spectral-ffn} then changes \emph{all} tokens.
For completeness, the conditioning and shrinkage identities are
\begin{equation}
 \kappa_2(\Phi_\rho(Z))
 =\frac{\lambda_{\max}+\rho\mu}{\lambda_{\min}+\rho\mu}
 \leq \frac{\lambda_{\max}}{\lambda_{\min}},
 \qquad \mu=\frac{\operatorname{tr}(Z)}{d}.
\end{equation}
\begin{equation}
 \frac{\Phi_\rho(Z)}{1+\rho}
 =\frac{1}{1+\rho}Z+\frac{\rho}{1+\rho}\mu I_d.
\end{equation}
The normalization in the second identity explains the relation to
isotropic shrinkage; it is not an extra operation performed by the
network. Initial values are $\alpha=0.2$, $\rho=0.1$, $b_i=0$,
and $\beta=2.2$.

\subsection{Training and Data Splits}
\label{app:protocol}

\begin{table}[htbp]
\centering
\caption{Fixed training recipes for SPD-MetaFormer and its paired ablations.}
\label{tab:training-details}
\small
\begin{tabular}{@{}lccc@{}}
\toprule
 & MI & SSVEP & ERN \\
\midrule
Optimizer & AdamW & Adam & AdamW \\
Learning rate & $2\times10^{-3}$ & $5\times10^{-3}$ & $5\times10^{-3}$ \\
Weight decay & $10^{-4}$ & $10^{-1}$ & $10^{-4}$ \\
Batch size & 128 & 64 & 64 \\
Epochs & 350 & 180 & 130 \\
Learning-rate schedule & cosine & constant & cosine \\
Gradient norm limit & 5 & none & 5 \\
Cross-entropy & unweighted & unweighted & class-weighted \\
Selection & min. val. loss & min. val. loss & min. val. loss \\
\bottomrule
\end{tabular}
\end{table}

Regularization applies to all trainable parameters, including the scalar
gate, spectral coefficient parameter, and readout parameters; these are
not excluded from weight decay. The SSVEP recipe therefore includes
Adam's coupled $L_2$ penalty of $0.1$.
No hyperparameters are retuned separately for the ablation configurations.

For MI, the preprocessed training and test sessions contain 288 trials
each. Signals are cropped to indices $[124,562)$, giving 438 samples.
Sorting training trials by class and reserving the last nine trials
within each class gives 252 training and 36 validation trials; the
separate evaluation session supplies 288 test trials.
For SSVEP and ERN, we follow the session-wise MAtt protocol \citep{pan2022matt}. The first four sessions are used for model development.
One of them is reserved for validation, and the remaining session is used for
testing.
ERN subjects are 2, 6, 7, 11, 12, 13, 14, 16, 17, 18, 20, 21, 22, 23,
24, and 26. Labels in the held-out test portion are not used for
checkpoint selection.

Writing $q_{sr}$ for the test metric of subject $s$ and run $r$, the
EEG scores report $\bar q\pm\overline{\mathrm{SD}}_{\rm run}$:
\begin{equation}
 \bar q_s=\frac{1}{R}\sum_{r=1}^{R}q_{sr},\qquad
 \bar q=\frac{1}{N}\sum_{s=1}^{N}\bar q_s.
\end{equation}
\begin{equation}
 \overline{\mathrm{SD}}_{\rm run}
 =\frac{1}{N}\sum_{s=1}^{N}
 \sqrt{\frac{1}{R-1}\sum_{r=1}^{R}(q_{sr}-\bar q_s)^2}.
\end{equation}
Here $R=10$, except that the temperature study uses $R=3$. This summarizes
training-run variability within subjects and defines both score error
bars and the entropy SD in Table~\ref{tab:entropy}. For paired comparisons, subjects remain the
sampling unit: the same averaging is applied to within-subject
differences before subjects are resampled, using 10,000 bootstrap draws.
Subject-level predictions for the published baselines are unavailable,
so Table~\ref{tab:benchmark} does not report paired statistical tests.

\subsection{Ablation Definitions and Parameter Counts}
\label{app:ablation-details}

The Euclidean control uses $D=d(d+1)/2$ log-vector features per token.
Its mixer averages all pre-normalized tokens, including the summary,
and adds this common vector to every token; a shared pre-normalized $D\!\to\!4D\!\to\!D$ GELU MLP
with an additive residual then transforms all tokens. The readout
averages the resulting token vectors. New classifiers and the Euclidean
MLP are initialized inside isolated RNG contexts, preserving the
shared initialization and the external RNG stream of the paired full
model. Concatenation changes classifier width, so it is a readout
design comparison rather than a parameter-matched control.

\begin{table}[htbp]
\centering
\caption{Trainable parameter counts for the configurations reported in the main ablation tables.}
\small
\begin{tabular}{@{}lrrr@{}}
\toprule
Configuration & MI & SSVEP & ERN \\
\midrule
Convolutional features only & 41,014 & 78,375 & 90,329 \\
SPD tangent readout & 8,114 & 73,725 & 75,289 \\
Euclidean MetaFormer & 122,954 & 185,805 & 85,765 \\
Memory-only geometric readout & 6,677 & 69,532 & 75,076 \\
Concatenation readout & 8,791 & 74,402 & 75,483 \\
Arithmetic pooling & 7,356 & 70,211 & 75,272 \\
Summary-only readout & 7,351 & 70,202 & 75,267 \\
Full SPD-MetaFormer & 7,356 & 70,211 & 75,272 \\
\bottomrule
\end{tabular}
\end{table}

The main-text comparisons describe selected structural and readout
questions; they are exploratory ablations, not a preregistered
confirmatory test of all architectural choices.

\subsection{Matched Uniform Controls for MAtt and GBWAtt}
\label{app:uniform-protocol}

The attention intervention uses ten seeds per subject, with 90, 110,
and 160 runs per model/condition on MI, SSVEP, and ERN. Each learned
model is compared with its own uniform counterpart, with identical
subject/seed keys, splits, learning rate, regularization, and epoch
budget. The GBWAtt comparisons use $\theta=1.5$ for both learned and
uniform variants.

Training settings for these matched comparisons differ from the
SPD-MetaFormer recipe. MAtt uses its mixed Adam/Stiefel optimizer with
learning rates $5\times10^{-4}$, $5\times10^{-3}$, and $10^{-2}$ for
MI, SSVEP, and ERN. GBWAtt uses RiemannianAdam with learning rates
$2.5\times10^{-3}$, $10^{-3}$, and $5\times10^{-4}$.
Both use weight decay $0.1$, batch sizes 128/64/32, and epoch budgets
350/180/130 in the same dataset order. MAtt batch sizes are the
implementation defaults; the experiment records do not contain per-run
batch sizes, so these values could not be independently verified.

MI uses the training-session holdout and separate test session described
above. For SSVEP, both architectures use the split of the released MAtt
and GBWAtt data loaders: sessions 1--3 (trials 1--300) for training,
session 4 (trials 301--400) for validation, and session 5 (trials 401--500)
for testing. For ERN, both architectures use the session-wise split
adopted by MAtt, with one of the first four sessions reserved for validation.
Within each architecture, the learned and uniform variants use identical
splits. These runs support the \emph{within-model} attention intervention.
Their scores are therefore reported separately from the published
benchmark comparison.

The released-structure GBWAtt implementation fixes its geometric matrix
parameter to the identity and aggregates transformed values as
\begin{equation}
 R_i=\left(\sum_{j=1}^{m}A_{ij}V_j^{1/2}\right)^2,
\end{equation}
with the same power transformations in the learned and uniform arms.
This is the implemented square-root-coordinate mean, not an assertion
that it equals the exact Bures--Wasserstein barycenter for arbitrary
noncommuting matrices. MAtt retains its LE mean. Thus, the intervention
changes the allocation weights without replacing either model's
aggregation rule.

Entropy is measured on the full test set after restoring the
minimum-validation-loss checkpoint. It is averaged first over query
rows and test trials, then over the ten seeds and subjects. Recorded
float32 uniform entropies agree with the analytical $\log m$ to within
$2\times10^{-6}$; Table~\ref{tab:entropy} shows the analytical reference.
The reported entropy comparisons concern these checkpoints and short
token sequences, not the entire training trajectory or all possible
attention architectures.

\clearpage
\section{Temperature Scaling in Manifold Attention}
\label{app:temperature}

We examine whether adjusting the softmax temperature can recover a
predictive benefit from learned attention. Keeping each model's score
$s_{ij}$ and aggregation geometry unchanged, we use
$A_{ij}^{(\tau)}=\operatorname{softmax}_{j}(s_{ij}/\tau)$.
For both MAtt and GBWAtt, we compare a retrained uniform control,
seven fixed temperatures
$\tau\in\{1,0.3,0.1,0.03,0.01,0.003,0.001\}$, and a learnable
$\log\tau$ initialized at $\tau=1$ or $0.01$.
Each condition is trained from scratch on all 9/11/16 MI/SSVEP/ERN
subjects with seeds 0--2, yielding 2,160 runs with matched initialization
and minibatch order.

Training recipes follow Appendix~\ref{app:uniform-protocol}, with
$\theta_{\rm GBW}=1.5$. For learnable temperature, the learning rate is
$0.01$, weight decay is zero, and $\tau$ is constrained to $[10^{-4},10]$.
All splits are disjoint: MI retains the split in
Appendix~\ref{app:protocol}, while SSVEP and ERN use consecutive
training/validation/test blocks of 300/100/100 and 180/60/100 trials,
respectively. Validation cross-entropy selects the checkpoint and,
within each subject/seed, the fixed temperature or learnable-temperature
initialization. Results average seeds within subjects before averaging
across subjects.

\begin{table}[htbp]
\centering
\caption{Temperature controls with three seeds. Default uses $\tau=1$; tuned and learnable settings are selected by validation loss. Scores are accuracy (MI/SSVEP) or AUC (ERN), multiplied by 100. $\Delta$ is tuned minus uniform, with a paired subject-bootstrap 95\% interval.}
\label{tab:temperature}
\setlength{\tabcolsep}{4pt}
\begin{tabular}{@{}lccccc@{}}
\toprule
Model / dataset & Default & Uniform & Tuned & Learnable & $\Delta$ [95\% interval] \\
\midrule
MAtt / MI & $74.74$ & $75.04$ & $74.09$ & $73.34$ & $-0.95\;[-2.40, +0.31]$ \\
MAtt / SSVEP & $64.39$ & $65.52$ & $66.27$ & $64.58$ & $+0.76\;[-0.36, +1.88]$ \\
MAtt / ERN & $71.35$ & $72.36$ & $73.65$ & $71.28$ & $+1.28\;[-1.78, +4.53]$ \\
GBWAtt / MI & $75.37$ & $75.18$ & $74.81$ & $73.33$ & $-0.37\;[-1.45, +0.54]$ \\
GBWAtt / SSVEP & $64.24$ & $63.55$ & $64.00$ & $63.97$ & $+0.45\;[-0.52, +1.36]$ \\
GBWAtt / ERN & $75.94$ & $75.21$ & $75.66$ & $74.52$ & $+0.45\;[-1.60, +2.92]$ \\
\bottomrule
\end{tabular}
\end{table}

\begin{figure}[!htbp]
\centering
\includegraphics[width=\linewidth]{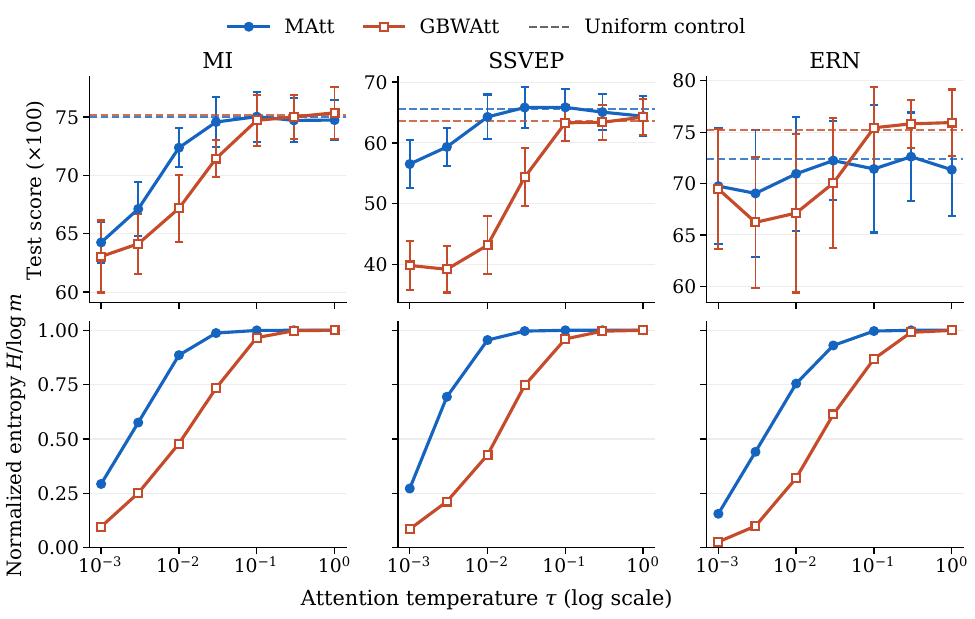}
\caption{Temperature, predictive performance, and allocation entropy after
full retraining. Top: mean test scores with the within-subject sample SD
over three runs, averaged over subjects; dashed lines
show each model's retrained uniform control. Scores are accuracy
(MI/SSVEP) or AUC (ERN), multiplied by 100, with separate vertical scales
across datasets. Bottom: normalized attention entropy, where one denotes
uniform allocation. All points average three seeds within subjects.}
\label{fig:temperature-curves}
\end{figure}

\begin{figure}[!htbp]
\centering
\includegraphics[width=\linewidth]{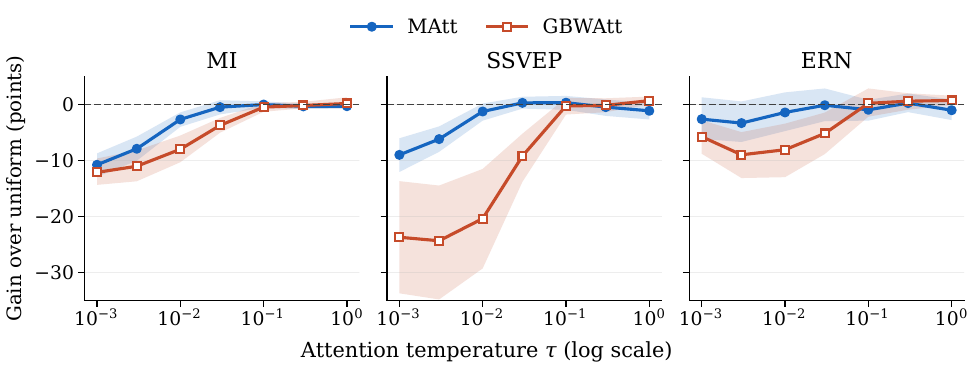}
\caption{Performance differences from retrained uniform mixing at each
fixed temperature. Positive values favor learned weighting. Shading shows
pointwise paired subject-bootstrap 95\% intervals after averaging seeds
within subjects. The dashed line denotes zero difference, and the
vertical scale is shared across datasets.}
\label{fig:temperature-effects}
\end{figure}

Lower temperatures substantially sharpen attention
(Figure~\ref{fig:temperature-curves}): at $\tau=0.001$,
normalized entropy ranges from $0.027$ to $0.293$ across the six settings,
but mean performance drops by $2.61$--$23.73$ points relative to uniform
(Figure~\ref{fig:temperature-effects}).
After validation selection, fixed temperatures change performance by
$-0.95$ to $+1.28$ points, with all paired intervals including zero
(Table~\ref{tab:temperature}). Learnable temperature likewise provides
no consistent improvement. Thus, despite gains over the default
temperature in some settings, changing temperature alone does not
reliably recover a predictive advantage over uniform geometric mixing
in these experiments.

\clearpage

\clearpage
\section{Functional Connectivity: ABIDE-I}
\label{app:fmri}

We evaluate SPD-MetaFormer on ABIDE-I autism classification to assess its
applicability to fMRI connectivity beyond EEG, keeping the backbone unchanged.

\paragraph{Data and tokens.}
The CPAC-preprocessed ABIDE-I release \citep{dimartino2014abide} of the
Preprocessed Connectomes Project \citep{craddock2013neuro} contains 1,035
subjects (505 with autism, 530 controls) from 20 sites. We use its
band-pass-filtered and global-signal-regressed CC200 time series
\citep{craddock2012whole}.
Four evenly spaced windows of $\operatorname{round}(T/2)$ frames, spanning
each $T$-frame scan from its first to last frame, provide four memory tokens;
the full scan provides the summary token. Each token is a shrunk Pearson
correlation matrix $0.3R+0.7I_{200}$. The entry map
(Equation~\ref{eq:input-bimap}) projects these tokens to $d=32$, without
a convolutional frontend. The resulting model has 10,538 parameters.

\paragraph{Protocol.}
All methods share ten 80/10/10 subject splits (828/103/104 subjects;
seeds 0--9), stratified by site and diagnosis. SPD-MetaFormer uses a fixed
AdamW recipe: learning rate $2\times10^{-3}$, weight decay $10^{-4}$,
batch size 32, class-weighted cross-entropy, gradient-norm clipping at 5,
and cosine decay over at most 20 epochs. The best validation-AUC checkpoint
is retained, with early stopping after five unimproved epochs. AUC uses
positive-class probabilities; accuracy and balanced accuracy use the
threshold in $\{0.20,0.21,\ldots,0.80\}$ maximizing validation accuracy.
Each test split is evaluated once. We report mean $\pm$ population SD
($\mathrm{ddof}=0$) across the ten splits.

\paragraph{Baselines.}
Euclidean baselines use the unshrunk full-scan correlation $R$: MLP and
$L_2$ logistic regression on its upper triangle, BrainNetCNN
\citep{kawahara2017brainnetcnn}, and a two-layer GCN on thresholded $R$
\citep{kipf2017gcn}. SPDNet-BN \citep{brooks2019riemannian} uses the shrunk
full-scan matrix with BiMap widths 100/50/30. MAtt \citep{pan2022matt} uses
the same five tokens as SPD-MetaFormer and $d_{\mathrm{out}}=32$, with
divided-difference spectral derivatives and a validation-selected AdamW
recipe. All model and threshold selection uses validation data only.

\begin{table}[htbp]
\centering
\caption{ABIDE-I (CC200) test performance: mean $\pm$ population SD
($\times100$) over ten shared subject splits (seeds 0--9).
SPD-MetaFormer and MAtt use $d=32$. Bold marks the largest mean in each column.}
\label{tab:abide}
\setlength{\tabcolsep}{5pt}
\begin{tabular}{@{}lccc@{}}
\toprule
Method & AUC & Accuracy & Balanced acc. \\
\midrule
\multicolumn{4}{@{}l}{\emph{Euclidean, full-scan correlation $R$}} \\
MLP on $\operatorname{vec}(R)$ & $78.65 \pm 2.67$ & $69.23 \pm 2.65$ & $69.16 \pm 2.48$ \\
$L_2$ logistic regression on $\operatorname{vec}(R)$ & $78.48 \pm 3.37$ & $71.25 \pm 3.11$ & $71.22 \pm 3.32$ \\
BrainNetCNN \citep{kawahara2017brainnetcnn} & $76.66 \pm 4.59$ & $69.04 \pm 4.07$ & $69.87 \pm 4.89$ \\
GCN on thresholded $R$ \citep{kipf2017gcn} & $72.65 \pm 3.65$ & $63.85 \pm 3.73$ & $65.40 \pm 3.04$ \\
\addlinespace
\multicolumn{4}{@{}l}{\emph{Riemannian}} \\
SPDNet-BN \citep{brooks2019riemannian} & $78.90 \pm 3.05$ & $71.15 \pm 3.30$ & $70.85 \pm 3.37$ \\
MAtt \citep{pan2022matt}, $d_{\mathrm{out}}=32$ & $77.24 \pm 3.61$ & $68.37 \pm 4.00$ & $68.18 \pm 4.52$ \\
\midrule
SPD-MetaFormer, $d=32$ & $\mathbf{79.14} \pm 3.53$ & $\mathbf{71.83} \pm 4.67$ & $\mathbf{71.79} \pm 4.66$ \\
\bottomrule
\end{tabular}
\end{table}

\paragraph{Results.}
SPD-MetaFormer achieves the highest mean AUC ($79.14$), accuracy ($71.83$),
and balanced accuracy ($71.79$) among the methods in
Table~\ref{tab:abide}. At $d=32$, its mean AUC exceeds MAtt by $1.90$
points while using approximately $57\%$ fewer parameters (10,538 versus
24,482). These results support the applicability of the same compact,
attention-free SPD backbone to both EEG covariance and fMRI connectivity.

\clearpage
\section{Spectral FFN Improves Robustness to Token Granularity}
\label{app:ffn-granularity}

Our spectral FFN $\Phi_\rho$ (Section~\ref{sec:spectral-ffn}) adds a
trace-proportional isotropic correction to each covariance token using
one learned scalar. We test its contribution by shortening the windows
used to estimate memory tokens, making covariance estimation more
challenging.

\paragraph{Setup.}
We vary the number of temporal windows $K$ ($3K$ memory tokens for
ERN's three channel banks) and train with or without the FFN, replacing
it by the identity in the latter case. The summary, frontend, splits,
training recipe, and checkpoint selection remain fixed. All 9/11/16
MI/SSVEP/ERN subjects and seeds 0--9 are included: 4,400 runs in total.
The main-model settings are $K=3/7/1$; the pre-specified comparison
settings are $K=18/4/6$, respectively. Statistics follow
Appendix~\ref{app:protocol}, with unadjusted paired $t$-tests over subjects.

\begin{figure}[htbp]
\centering
\includegraphics[width=\linewidth]{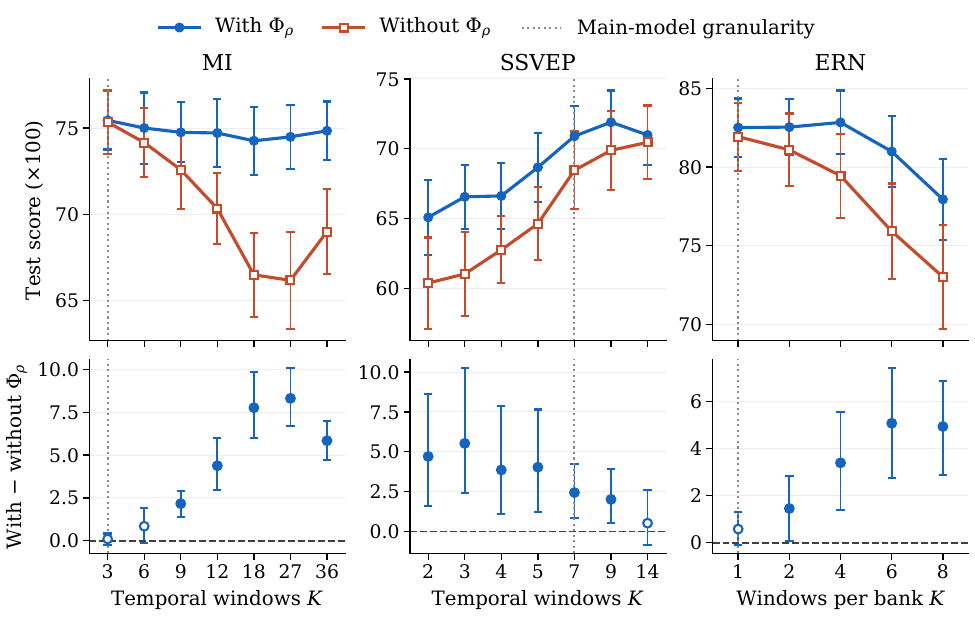}
\caption{\textbf{Contribution of the spectral FFN across token granularities.}
Top: accuracy (MI/SSVEP) or AUC (ERN), $\times100$, with the within-subject
sample SD over ten runs, averaged over subjects.
Bottom: FFN gain with subject-bootstrap 95\% intervals; filled markers
indicate intervals excluding zero. Ten seeds are averaged within each
subject. Dotted lines mark the main-model settings; these single-threaded
reruns differ from Table~\ref{tab:benchmark} by at most 0.11 points.}
\label{fig:ffn-granularity}
\end{figure}

\paragraph{Robustness from one parameter.}
The FFN substantially reduces the performance loss from finer MI and ERN
tokens (Figure~\ref{fig:ffn-granularity}). At the pre-specified settings,
it improves MI accuracy by $7.76$ points (95\% interval $[5.99,9.82]$)
and ERN AUC by $5.06$ points ($[2.76,7.40]$), benefiting 9/9 and 15/16
subjects, respectively (both $p<0.002$). Across all seven MI
granularities, accuracy stays within $74.3$--$75.5\%$: a range of only
$1.2$ points, versus $9.2$ without the FFN. SSVEP also benefits at its
main setting ($+2.45$ points, $[0.87,4.22]$, $p=0.021$) and at the
pre-specified $K=4$ ($+3.86$, $[1.12,7.89]$, $p=0.073$).
At the coarser main settings of MI and ERN, the gains are small
($+0.10$ and $+0.58$; both intervals include zero), locating the
clearest benefit in the more demanding tokenizations.

\paragraph{Spectral conditioning.}
At the selected checkpoints, the geometric-mean condition number of
memory tokens after the FFN stays below $10^{2.3}$ across all settings,
whereas it reaches $10^{5.9}$ without the FFN. The learned correction
remains mild ($0.04\le\rho\le0.07$). Together, these results support
the FFN as a compact spectral conditioner that improves robustness to
token granularity with a single learned parameter.

\end{document}